\documentclass[11pt,a4paper]{article}
\usepackage{amsmath}
\usepackage{txfonts}
\usepackage{microtype}
\usepackage[hyperref]{emnlp2020}
\usepackage{latexsym}

\usepackage{booktabs}
\usepackage{dirtytalk}
\usepackage{multirow}
\usepackage{simplex}

\aclfinalcopy 
\def\aclpaperid{***} 

\title{Detecting Objectifying Language in Online Professor Reviews}
\author{Angie Waller and Kyle Gorman \\
Graduate Center, City University of New York}

\date{}
\usepackage{graphicx}
\begin{document}

\maketitle

\begin{abstract}
Student reviews often make reference to professors' physical appearances. Until recently RateMyProfessors.com, the website of this study's focus, used a design feature to encourage a \say{hot or not} rating of college professors. In the wake of recent \#MeToo and \#TimesUp movements, social awareness of the inappropriateness of these reviews has grown; however, objectifying comments remain and continue to be posted in this online context. We describe two supervised text classifiers for detecting objectifying commentary in professor reviews. We then ensemble these classifiers and use the resulting model to track objectifying commentary at scale. We measure correlations between objectifying commentary, changes to the review website interface, and teacher gender across a ten-year period.
\end{abstract}
\vspace{1mm}

\section{Introduction}

Natural language processing techniques have long been used to study subjectivity and sentiment in media and product reviews.
In this study, we employ these technologies to study
objectifying language in reviews of professors
using archival data from RateMyProfessors.com (RMP).
Detecting such language is difficult
because it is somewhat rare,
making up a small part of a small proportion of reviews \citep{davison},
and references to physical appearance show
enormous linguistic variation (discussed in Section \ref{reviewex}),
making them difficult to detect accurately using simple text features.

This study provides insights into bias in professor reviews
and their interaction with the design of the web user interface.
We propose two models---a chunk tagger and a document classifier---used to build an ensemble to detect objectifying reviews at scale.
This approach could be applied to many other domains where noisy user-generated reviews may contain harassment or exhibit harm.

We focus on the RMP website because it has been active for over twenty years, giving us ample data to study trends across time. The website has long been associated with students commenting on their professors' appearances \citep{Voice} and has been the subject of many prior studies on bias in course reviews. Recent changes to the website interface allow us to consider how text reviews may have been influenced by its design feature for rating professor \say{hotness}.

\subsection{Prior work} 
We look to previous work on bias in professor reviews, effects of interface design on internet discourse, and detecting subjectivity and opinions in online reviews. 

\subsubsection{Bias and student reviews}
Prior studies address bias among students’ reviews of teachers. \citet{freng} ﬁnd a positive correlation between \say{hotness} and quality scores of professors on RMP, accounting for 8\% of variance.
\citet{chang-mckeown-2019-automatically} report gendered differences in students' descriptions of computer science professors on RMP which is also reflected in visualizations by \citet{schmidt} showing that
words like \emph{genius} are more frequently attributed to male professors and words like \emph{nurturing} to female professors.
This is supported in work by \citet{effectiveness} and \citet{BORING201727} where in-class reviews show higher ratings for leadership skills among male professors and \say{being warm} among female professors. Noting that perceptions of \say{easiness} predict overall ratings, \citet{davison} recommend an RMP interface change replacing the site's \say{easiness} rating with better-defined terms such as \say{amount learned}.

\subsubsection{Interface design and online discourse}

The interaction between interface design and online discourse is a central focus in computer-mediated discourse analysis \citep{herring_2004,herring2} and critical technocultural discourse analysis \citep{CTDA}.
Both consider not only how people express themselves in online environments, but also how elements like interface design of a website shape people into \say{users}, affecting how they express themselves. 
Because we are interested in the relationship between attractiveness commentary and interface design, we constrain this study to the RMP website and its interface elements, including its professor rating form (Appendix, \autoref{fig:webform}), featuring the \say{hot or not} chili pepper rating. 

Interface design is also considered quantitatively, and at scale, in company-led user experience studies. For example, Facebook found that by curating users' News Feeds to positive or negative posts, they influenced the emotional tenor of the users' own posts \citep{Kramer8788}. NextDoor, a popular neighborhood classifieds website, made their web form for reporting suspicious activity more detailed and inadvertently arduous, successfully decreasing suspicious activity posts and therefore decreasing posts with racial profiling \citep{hempel_for_2017}. In an effort to combat online harassment, Twitter introduced interface elements to warn users before posting tweets with inflammatory language \citep{twitter_warning}. These interventions suggest that small interface changes may produce measurable effects in online discourse.

\subsubsection{Subjectivity in workplace reviews}

Like all genres of review, professor reviews interweave subjective and objective statements (e.g., \say{the class was poorly attended}).
We consider commentary on a professor's physical attractiveness, which we refer to as objectifying or attractiveness commentary, to be subjective content.

\citet{wiebeetalacl01wkshop} discuss a method of labeling spans of subjective text within news corpora so that opinion phrases, even those that occur infrequently, can be detected using collocation clues. To determine review sentiment, \citet{10.3115/1218955.1218990} 
automatically segment movie reviews into subjective and objective portions, discarding the objective portions before attempting to determine the overall sentiment. 
Here we are also interested primarily in a subjective portion of reviews,
but whereas subjectivity is an expected feature in other genres of reviews,
comments about a professor's \say{hotness} may constitute workplace harassment \citep{nochili} among other harms. To our knowledge, this is the first study to target objectifying commentary in professor reviews and its relationship to website design.

\subsection{Our approach}

Our classification scheme is tailored for a low-resource setting with a limited amount of labeled data.
The goal is to construct classifiers which achieve sufficient accuracy to allow extrapolation to a much larger set of unlabeled reviews. 
To achieve our goal of analyzing large-scale trends, we train two models for identifying objectifying commentary in RMP reviews: (1) a token-level chunk tagger similar to those used for named entity recognition; and (2) a review-level text classiﬁer similar to those used for document classification. Unlike the chunk tagger, the document classifier can take into account more variety in features and account for attractiveness commentary that occurs multiple times in a document. Multiple spans can \say{gang up},
allowing them to be more easily detected at document level. In contrast, the chunk tagger considers objectifying language as a highly-local phenomena and is therefore more able to detect attractiveness commentary in the context of longer reviews covering a range of topics.

We then build ensembles of these models.
We anticipate that ensembling will be useful because we hypothesize that
the two classifiers' patterns of errors will be only weakly correlated \citep{van-halteren-etal-1998-improving},
and because labeled data is limited,
high-variance,
and class-imbalanced \citep{brill-wu-1998-classifier}
for this task.

\section{Data}
For this study, anonymous RMP reviews of professors were scraped on two occasions.%
\footnote{
    Scraping was seeded using a list of professors and their chili pepper scores (\url{http://morph.io/chrisguags}).
}
The first scrape, in July 2018, paired textual data with the professor's \say{hotness} rating, defined by the number of times a student rated the professor as \say{hot} minus the number times they rated them as \say{not hot} \citep{attractive}.
In the web interface, the names of professors receiving positive attractiveness scores are marked with a chili pepper emoji (see Appendix, \autoref{fig:prof_listing}). 
The second scrape, in August 2019, targeted a broader set of regions and schools.
Test data was drawn from this latter data set,
which was also used for trend analysis.

By this latter date, the chili pepper emoji had been removed from the website in response to public criticism \citep{byechili2},
so it was no longer possible to extract hotness scores.
In addition to text, both scrapes also collected the names of professors,
student-reported quality and difficulty scores (averaged by professor, on a five-point scale),
subject area, and the name of the school.
See \autoref{table:d1colleges} and \autoref{table:d2colleges} in the Appendix for the full list of schools.

\subsection{Defining objectifying language}

We define objectifying or attractiveness commentary as reviews that describe a professor's physical appearance, demeanor, clothing style, or resemblance. In contrast to prior work \citep[e.g.,][]{attractive}, we also include language disparaging a professor's appearance. Although previous work has considered objectifying comments in limited RMP datasets \citep{davison, kindred}, there are no previous annotation guidelines to follow for labeling these expressions. \citet{kindred} find out of 788 RMP ratings in their sample, only 3.6\% describe teacher attractiveness. Given the low frequency of these reviews and their informal qualities, creating instructions that cover attractiveness commentary in all of its variations is not possible. We acknowledge some reviews like ones described in Section \ref{fuzz} will be more subjective than others.

\subsection{Review characteristics}\label{reviewex}

RMP reviews contain stylistic flourishes common to online discourse:
slang and non-standard language,
typographical errors,
expressive punctuation and capitalization,
and emoticons. The examples below are fragments 
from 30-to-50-word reviews representing attractiveness commentary.
See \autoref{fig:GhostPeppers} in the Appendix for additional examples
in screen-capture format.

\begin{itemize}

\item \emph{Everyone LOOOOOVES sexy Jeff!}
\item \emph{\ldots he doesn't assume students understand complex stuff like other math teachers do. Plus, hello, HOT!} 
\item \emph{He's also pretty cute which helps. :)}
\item \emph{\ldots when he talked about vector space he almost saw my O-face.}
\end{itemize}

\subsection{Fuzzy samples} \label{fuzz}
This section describes reviews that pose challenges in labeling attractiveness spans and the process behind how distinctions are made. Annotators were instructed that, when in doubt, reviews that imply romantic interest, or lack thereof, are considered objectifying.
\paragraph{Flirtation but no attractiveness commentary}
Examples where the review may be flirtatious but not directly describing professor appearance present a grey area.

\begin{itemize}
\item \emph{Damn, I love that man. \hfill \small{\textbf{None}}}
\end{itemize}
Referring to a professor as \say{that man} borders on objectifying, but without additional context it is not considered attractiveness commentary.

\begin{itemize}
\item{\textit{I love him so much, I would totally  marry him  if I could.}\hfill \small{\emph{\textbf{Obj.}}}}
\end{itemize}

\noindent
However, we consider references to marriage or dating the professor like the above example to be objectifying commentary.
We decide this because the element of fantasy in samples like these is taken to be indicative of an attraction to the professor.

\begin{itemize}
\item{\textit{He is a math god!} \hfill \small{\textbf{None}} }
\end{itemize}
Reviews that compare the  professor to a deity are also difficult to distinguish. If the focus could be the professor's expertise, the review is not considered attractiveness commentary. 

\paragraph{Accents} 
The most common challenging examples refer to the professor’s voice or accent. These types of reviews primarily fall into two categories: (1) the accent is sexy, charming, or appealing; and (2) denoting professors who are non-native English speakers described as difficult to understand. Reviews in the latter category can be considered denigrating of the professor but are not necessarily attractiveness commentary. We consider the intent of the student. If the comment is personally derogatory, such as \say{horrible accent}, it is considered objectifying. The following examples illustrate these distinctions:
\begin{itemize}
\item{\textit{And he’s British, such a  charmer! Love his accent!}\hfill \small{\emph{\textbf{Obj.}}}}
\item{\textit{He has the cutest accent.}\hfill \small{\emph{\textbf{Obj.}}} }
\item{\textit{His accent was difficult to understand.}\hfill \small{\textbf{None}}}
\end{itemize}

\begin{table*}[t]

\centering

\begin{tabular}{l | c  c  c  c  c  c  c  c  c  c }
\toprule
  & He & is & CUT & for & a & Stanford & professor \\
\midrule
\small\verb|word-lower| & \small\verb|he| & \small\verb|is| & \small\verb|cut| & \small\verb|for| & \small\verb|a| & \small\verb|stanford| & \small\verb|professor| \\
\small\verb|lemma| & \small\verb|he| & \small\verb|is| & \small\verb|cut| & \small\verb|for| & \small\verb|a| & \small\verb|stanford| & \small\verb|professor|
\\
\small\verb|pos| & \small\verb|PRON| & \small\verb|AUX| & \small\verb|NOUN| & \small\verb|NOUN| & \small\verb|DET| & \small\verb|PROPN| & \small\verb|NOUN|
\\
\small\verb|has-hot| & \small\verb|false| & \small\verb|false| & \small\verb|false| & \small\verb|false| & \small\verb|false| & \small\verb|false| & \small\verb|false|
\\
\small\verb|next-word| & \small\verb|is| & \small\verb|CUT| & \small\verb|for| & \small\verb|a| & \small\verb|stanford| & \small\verb|professor| & \small\verb|[END]|
\\
\small\verb|next-pos| & \small\verb|AUX| & \small\verb|NOUN| & \small\verb|ADP| & \small\verb|DET| & \small\verb|PROPN| & \small\verb|NOUN| & \small\verb|[END]|
\\
\small\verb|prev-word| & \small\verb|[START]| & \small\verb|He| & \small\verb|is| & \small\verb|cut| & \small\verb|for| & \small\verb|a| & \small\verb|stanford|
\\
\small\verb|prev-pos| & \small\verb|[START]| & \small\verb|PRON| & \small\verb|AUX| & \small\verb|NOUN| & \small\verb|ADP| & \small\verb|DET| & \small\verb|PROPN|
\\
\small\verb|prev-iob| & \small\verb|O| & \small\verb|O| & \small\verb|O| & \small\verb|B| & \small\verb|O| & \small\verb|O| & \small\verb|O|
\\
\small\verb|all-caps| & \small\verb|false| & \small\verb|false| & \small\verb|true| & \small\verb|false| & \small\verb|false| & \small\verb|false| & \small\verb|false|
\\
\small\verb|prev-all-caps| & \small\verb|false| & \small\verb|false| & \small\verb|false| & \small\verb|true| & \small\verb|false| & \small\verb|false| & \small\verb|false|
\\
\small\verb|next-all-caps| & \small\verb|false| & \small\verb|true| & \small\verb|false| & \small\verb|false| & \small\verb|false| & \small\verb|false| & \small\verb|false|
\\
\midrule
\end{tabular}
\caption{Example feature vector for chunk tagger using review snippet commenting on professor's physique.}
\label{tab:iob1}
\end{table*}

\section{Methods}

We implement classification techniques with unique strengths for capturing the qualities and contexts of objectifying comments. The first, a chunk tagger, represents a bottom-up strategy, whereas the second, a document classifier, uses top-down processing and a richer feature set.

\subsection{Chunk tagger}

Because discussion of a professor's attractiveness may only be a small portion of any given review, we annotate spans of tokens which refer to attractiveness.
These labels can then be automatically propagated from spans to the document level. That is, if a review contains any spans tagged as containing objectifying language, the whole review is labeled objectifying. We employ a chunk tagger customized to identify these spans within reviews. During preprocessing, labeled data is tagged for part of speech (POS) using the \texttt{spaCy} tagger \citep{spacy2}.
Text spans that refer to attractiveness are tagged using the CoNLL-2003 IOB format \citep{tjong-kim-sang-de-meulder-2003-introduction}.
The chunker is built using the \texttt{nltk.chunk} library \citep[ch.~7]{bird_natural_2009};
it uses a multinomial logistic regression classifier and a greedy left-to-right decoding strategy.

\paragraph{Attractiveness features}
In addition to token features, we develop a dictionary of words describing attractiveness (see Appendix, \autoref{table:dictionary});
these are matched using regular expressions so alternative spellings (e.g., \emph{hoooottt}, \emph{hotttttt}) are also captured.
See \autoref{tab:iob1} for an example token feature vector.

\subsection{Document classifier}
We also develop a model that can take advantage of features extracted from the entire review.
The document classifier is built using a linear-kernel support vector machine classifier
from  \texttt{sklearn} \citep{scikit-learn}.
The primary features used are term frequency-inverse document frequency weighted unigrams and bigrams.
Several other types of features, described below, are used to improve classifier accuracy.
\begin{table}[b]
    \centering
    \begin{tabular}{l r r r r}
    \toprule
         & Reviews &  Tokens & Words \\ 
    \midrule
    Labeled & 4,050 & 12,209 & 139,091 \\
    Unlabeled &358,970&71,700&15m\\
    \bottomrule
    \end{tabular}
    \caption{Summary statistics for datasets.}
    \label{tab:partitions}
\end{table}
\paragraph{Formality}
Impressionistically, RMP reviews that discuss teacher appearance tend to be less formal 
than those that focus on the quality of instruction.
To capture this distinction, we use features proposed by 
\citet{pavlick-tetreault-2016-empirical} 
to measure textual formality.
These include 
average word and sentence length,
the ratio of nouns to verbs,
and the proportion of words over 4 characters.
We also add one-hot features for the use of non-standard punctuation and capitalization.
Finally, we also extract features tracking the use of titles such as \emph{Dr.}, \emph{Professor}, \emph{Mrs.}, and \emph{Mr.}

\paragraph{Gender}
We extract professor gender 
by tracking third-person singular pronouns (e.g., \emph{he}, \emph{his}, \emph{she}, \emph{her}) in reviews;
gender-non-specific pronouns like \emph{they} and
neo-pronouns like \emph{ze} were not present in the labeled data and therefore not tracked.
We also do not track gender of the reviewers as all reviews are submitted anonymously.

\paragraph{Subjectivity}
\citet{davison} and \citet{Ritter} argue that student reviews largely follow a transactional consumerist discourse similar to customer service reviews.
We hypothesize that this would be reflected in the ratio of first-person to third-person pronouns; a greater proportion of first-person pronouns may indicate a review about personal opinions and feelings (\emph{consumerist}) rather than instruction. We also reuse the attractiveness dictionary regular expression patterns from the chunk tagger,
expanding this to include common idioms such as \emph{easy on the eyes} and \emph{good looking}.
Additionally, each review is scored for its sentiment and subjectivity using the \texttt{textblob}\footnote{\texttt{https://textblob.readthedocs.org}} sentiment classifier.

\paragraph{Style}
We consider features measuring the use of text properties characteristic of internet discourse,
including the use of emoticons, repeated exclamation points, and words in all uppercase letters.

\subsection{Feature ablation}
For the document classifier, a feature ablation study on the development data (Appendix, \autoref{table:featurestest}) shows accuracy scores rely on the custom dictionaries for \say{hotness} including pattern matching for idiomatic expressions. However, omitting formality and stylistic features does not impact performance.

\subsection{Ensembling}

After training the chunk tagger and document classifier on the labeled data, a simple document-level ensemble of these models is applied to the unlabeled data. Since there are only two weak classifiers available, we use two forms of voting: in ensemble 1 we consider reviews with disagreement as non-objectifying reviews; in ensemble 2, we completely discard reviews when the classifiers disagree to achieve higher accuracy.

\section{Results}

\subsection{Experiment setup}
The labeled data of 4,050 reviews is randomly split into training (80\%) and development sets (20\%),
the latter used for feature ablation (Appendix, \autoref{table:featurestest}).
During annotation, professors labeled \say{hot} were 
deliberately oversampled.
Review and token counts can be found in \autoref{tab:partitions}.

\begin{table}[t]
\centering
\begin{tabular}{lrr}
\toprule
\multirow{2}{*}{Chunk tagger} &
\multicolumn{2}{c}{Doc.~classifier} \\
\cmidrule{2-3}
                 & Targeted & None    \\
\midrule
Targeted         & 8,573    &   9,858 \\
None             & 4,295    & 336,242 \\
\bottomrule 
\end{tabular}
\caption{Confusion matrix for chunk tagger and document classifier models; Targeted: 
reviews which contain attractiveness commentary.}
\label{table:confusionmatrix}
\end{table}
\begin{table}[b]
\centering
\begin{tabular}{ l r r r r r }
\toprule
Classifier&{Prec.}&\multicolumn{1}{c}{Rec.}&\multicolumn{1}{c}{$F1$}&\multicolumn{1}{c}{Acc.}&\multicolumn{1}{c}{$\kappa$}\\
\midrule
Chunk tag.    & .42 & .21 & .28 & .89 & .23 \\
Doc. class. & .44 & .23 & .30 & .93 & .26 \\
\bottomrule
\end{tabular}
\caption{Weak classifier results.}
\label{table:doc_class}   
\end{table}

\begin{table*}[h!]

    \centering
    \begin{tabular}{p{10cm}p{1.5cm}p{1.5cm}}
    \toprule
        Review & \multicolumn{1}{c}{Chunk tagger} &  \multicolumn{1}{c}{Doc. class.}  \\ 
   \toprule
  \small  
\textit{Not a great teacher (in fact pretty awful) but she's looking GOOD.}
 & \multicolumn{1}{c}{FN} & \multicolumn{1}{c}{TP}  \\
    \midrule
    \small  
\textit{he is now bald, but he still has the look ;)}
 & \multicolumn{1}{c}{FN}& \multicolumn{1}{c}{TP} \\
  \midrule
    \small  
\textit{His classes are worthwhile because he's a good teacher, but mostly because he has the most awesome accent in the world. Rawr.}
 & \multicolumn{1}{c}{FN}& \multicolumn{1}{c}{TP} \\
     \midrule
     \small  
\textit{the WORST ****ING TEACHER EVER. WORST CLASS, WORST PERSON. NOT PROFESSIONAL AT ANYTHING, DOES NOT KNOW PHYSICS FROM THE HOLE IN HIS ASS. AVOID!}
 & \multicolumn{1}{c}{FP}& \multicolumn{1}{c}{TN} \\
  \midrule
     \small  
\textit{My experience with this professor was awful. He wasn't helpful and I ended up learning everything on my own without his help. I should have just stared at the wall rather than wasting my time in this class. He did not BUMP my grade up!}
 & \multicolumn{1}{c}{FP}& \multicolumn{1}{c}{TN} \\
     \midrule
     \small  
\textit{Probably the BEST Org Chem prof out of all the ones I've had. His slides are actually notes, not just pictures with lines on the side for you to write on. The exam is based onthe  notes, but you also need to read the book. Def didn't mind looking at him for 1 hour 25 mins either.}
 & \multicolumn{1}{c}{TP} & \multicolumn{1}{c}{FN}  \\
   \midrule
   \small  
\textit{not a bad prof. has a nice smile. class discussions were pretty interesting. grades are based on ur attendance, and ur blog entries (they are not hard, but be careful, cuz her way of grading is kinda picky). overall, not a hard class. kinda interesting. take it if u want, but if u cant stand reading don't. TONS of reading.}
 & \multicolumn{1}{c}{TP} & \multicolumn{1}{c}{FN} \\
 \midrule
   \small  
\textit{Jason's a fantastic section leader--some of the best classes I've had here  were in section for this class. Plus, he knows his stuff, is super eloquent, and kicks ass in suits (just sayin'). I will say that he can come off as cold and intimidating at first, but he actually cares and is really willing to help you.}
 & \multicolumn{1}{c}{TP} & \multicolumn{1}{c}{FN} \\
  \midrule
  \small  
\textit{I can't understand her heavy accent. I found her subject boring.}
 & \multicolumn{1}{c}{TN}& \multicolumn{1}{c}{FP} \\
 \midrule
  \small  
\textit{Jenny is an extraordinary professor- she truly cares about how you do in her class, and does her best to help you in whatever fashion she can.}
 & \multicolumn{1}{c}{TN}& \multicolumn{1}{c}{FP} \\
 \midrule
\small  
\textit{Very easy going. Knows what he is doing from lived experience. The powerpoints are very good. You can skip class and just follow along on the slides and get the idea of things (although probably not a good grade). Hot daughter.} & \multicolumn{1}{c}{FP} & \multicolumn{1}{c}{FP}  \\ 
 \midrule
 \small  
\textit{worst prof ever, and i really mean that. she's not even a professor, just some plant biologist hired as a lecturer. she is completely inept as a lab manager and universally hated by the students. oh, and very not hot} & \multicolumn{1}{c}{TP} & \multicolumn{1}{c}{TP}  \\ 
 
    \bottomrule
    \end{tabular}
    \caption{Examples with classifier disagreement; reviews have been modified to reflect their original format while protecting the identity of professors.}
    \label{tab:disagreementexample}
\end{table*}

\subsection{Annotation}


To estimate interannotator agreement,
a subset of the labeled data was independently 
labeled by a second annotator, a graduate student in linguistics, according to the authors' guidelines.
This gave a span-level Cohen's $\kappa~{=}~.785$
and a document-level $\kappa~{=}~.801$;
both correspond to \say{substantial} agreement according to the \citet{landis_1977} qualitative guidelines.

\subsection{Performance}

After applying the chunk tagger and document classifier
to the unlabeled data, we find the classifiers disagree on 4.1\% of the reviews (see Table \ref{table:confusionmatrix}).  This is roughly what one might expect given the overall low proportion of true positive samples. We determine accuracy by creating a test set from 600 of these reviews. This set includes reviews with classifier agreement on 150 documents predicted to contain,
and 150 documents predicted not to contain, objectifying language.
We also sample 300 reviews in which the chunk tagger and document classifier disagree.\footnote{We oversample from recent date ranges to better capture any new trends in reviews.}
These 600 samples are randomly sorted and then adjudicated by a human judge
to create the test set.

\begin{figure*}[t!]
\centering
\includegraphics[width=1\textwidth]{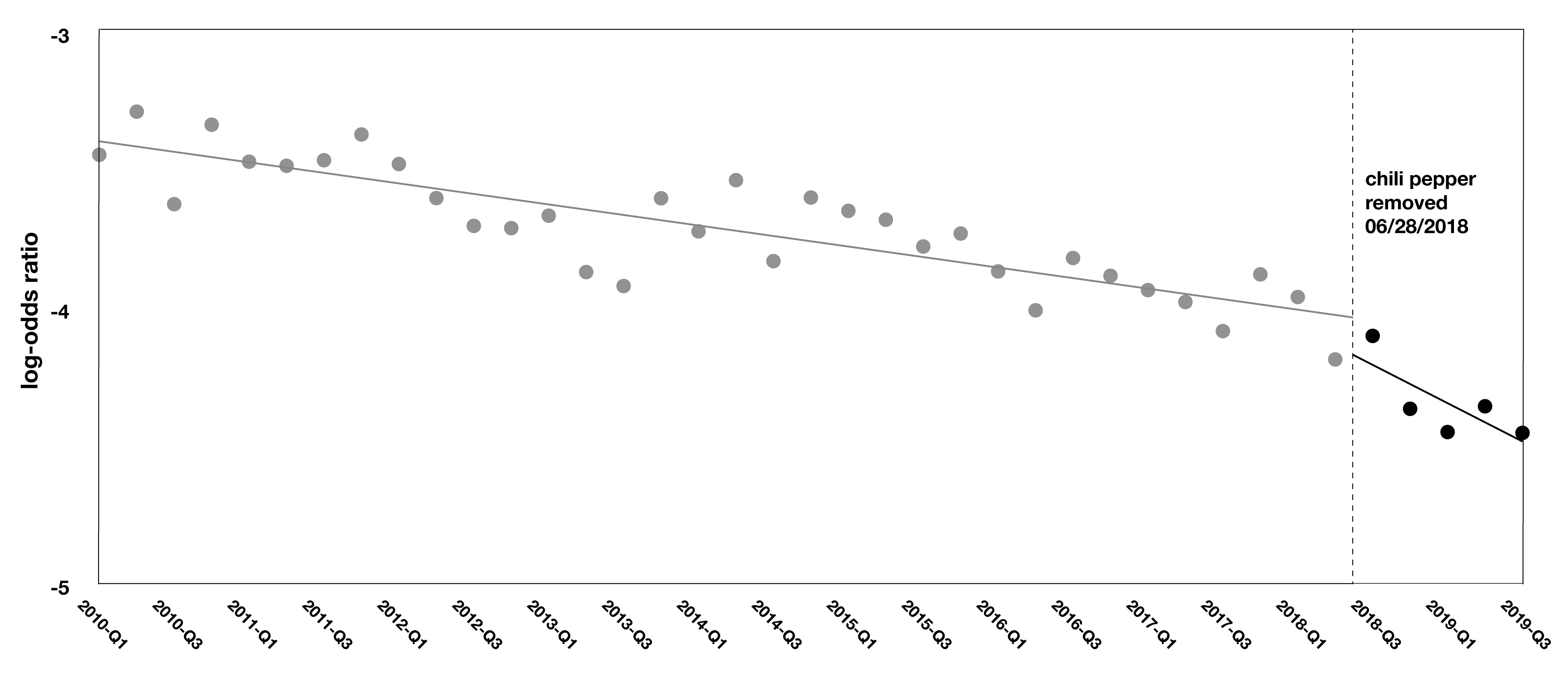}
\caption{Log-odds of attractiveness commentary in reviews from 2010 to August 2019.}
\label{fig:Longitude}
\end{figure*}

The results for the chunk tagger and document classifier are shown in  \autoref{table:doc_class}.
As can be seen, both classifiers have relatively high accuracy but significantly lower precision and recall. Table \ref{tab:disagreementexample} shows example reviews where the classifiers disagree. The chunk tagger performs better in reviews with higher word counts. In some cases, the chunk tagger avoids false positives of the document classifier where keywords from custom dictionaries appear but are in a context that is not objectifying. The document classifier performed well on lower word count reviews and where words from custom dictionaries and regular expression patterns are present. In \autoref{table:ensembles}, we show the same results for the two ensembles;
note that results for ensemble 2 do not include the 300 samples from the test data on which the two weak models disagree.

We see that both ensemble classifiers achieve greatly improved results compared to either the chunk tagger or the document classifier alone,
and as expected, error can be further reduced in ensemble 2 by discarding
data on which they disagree.\hfill \break
\hspace*{6mm}We conclude that ensemble methods are effective for detecting objectifying commentary in student reviews in the face of unbalanced data.
In what follows, the ensemble 2 classifier is used to 
to analyze trends in attractiveness commentary on 344,815 reviews.
\begin{table}[]
\centering
\begin{tabular}{ l r r r r r }
\toprule
Classifier&{Prec.}&\multicolumn{1}{c}{Rec.}&\multicolumn{1}{c}{$F1$}&\multicolumn{1}{c}{Acc.}&\multicolumn{1}{c}{$\kappa$}\\
\midrule
Ensemble 1 &.72&.44&.55&.93&.50\\
Ensemble 2 &.72&1.00&.84&.99&.83\\
\bottomrule
\end{tabular}
\caption{Ensemble classifier results.}
\label{table:ensembles}
\end{table}
\section{Analysis}
Building on previous RMP research studying bias in student reviews, we continue this inquiry focusing on how attractiveness commentary is distributed based on teacher gender, and quality and difficulty scores. We then focus on a logistic regression analysis using generalized estimating equations (GEE) to determine if there was a decrease in attractiveness commentary following the removal of the chili pepper feature from the web interface. 

\subsection{Teacher gender}
Our dataset contains 39.7\% female professors ($11,192$) compared to 60.2\% male professors ($16,967$). This proportion is similar to those found in U.S. higher education where women make up only 31\% of full-time faculty \citep{Kelly_2019}. Since this breakdown leads to more reviews for male professors overall, we consider each professor and whether or not they have at least one objectifying comment. In our dataset, 21.0\% of male professors have at least one attractiveness review compared to 18.4\% of female professors. We find in a chi-square test of independence that this difference is significant ($\chi^2 = 17.75$, $p <.01$).
In contrast, \citet{rosen} found that women were more likely to have the chili pepper rating (27.8\%) than men (22.7\%). We believe this difference could be attributed to the distinction between the low effort act of clicking \say{hot} on the review form versus actually writing commentary on the teacher's appearance. Also, unlike chili pepper ratings, our counts include reviews with negative commentary. 


\subsection{Logistic regression} \label{logreg}

We deploy logistic generalized estimating equations (GEE; \citealt{GEE10.2307/2336267}), an extension of the generalized linear modeling that takes into account the correlation between observations. A logistic GEE accommodates the unequal number of earlier observations across professors and conditions as well as the variation in review activity volume over quarterly time intervals. This is optimal for the noise in the dataset and allows utilization of the entire collection of reviews. The final model parameters are determined by the best goodness-of-fit score computed using the full log quasi-likelihood function. School size and tuition did not have significant outcomes in the results and were discarded.
The final model includes presence or absence of the chili pepper interface feature, teacher quality and difficulty scores, and professor gender. Time is input as an interval covariate by quarter, while chili pepper condition is a binary factor; final parameters and their outcomes are given in \autoref{tab:GEEModel}.

\begin{table*}[h!]
 \centering
    \begin{tabular}{l r r r r}
  \toprule
     & Estimate &   & &    \\
     & \small{(log-odds)} &  Std.~err. &Wald $\chi^2$&   $p(\chi^2)$ \\
    \midrule
         (Intercept) & $-3.111$  &   .143&    476.18&    $< .001$ \\
         pepperAbsent&  $-.428$  &  .136&   9.93&  .002\\
         timeInQuarters&  $-.020$  &  .002& 79.44& $< .001$ \\
         difficultyHigh& $-.075$  &  .022&  11.49&   $< .001$\\
         qualityHigh& $.051$  &  .026 &  3.76&   .053 \\
         genderFemale& $-.528$  &  .174 &  9.19&   .002 \\
         qualityHigh:genderFemale& $.097$  &  .043 & 5.09 &   .024 \\
           
         \bottomrule
    \end{tabular}
    \caption{GEE model parameter estimates with attractiveness commentary as dependent variable.
    The intercept represents pepperPresent, timeInQuarters = 0, difficultyLow, qualityLow, genderMale. $N = 344,815$.}
    \label{tab:GEEModel}
\end{table*}

\paragraph{Chili pepper and time interval}

First, we focus on our primary question concerning the proportion of objectifying comments and the removal of the chili pepper. We observe a downward trend over the time period prior to the interface change; however, the log-odds of attractiveness commentary after the chili pepper was removed on June 28, 2018 is lower than the time variable can account for alone (see \autoref{tab:GEEModel}). Our analysis finds a significant effect of time and condition (with vs.~without the chili pepper). These findings support our hypothesis: RMP's removal of the chili pepper coincides with a decline in reviews mentioning professor attractiveness.

\begin{figure*}
\centering
\includegraphics[width=0.99\textwidth]{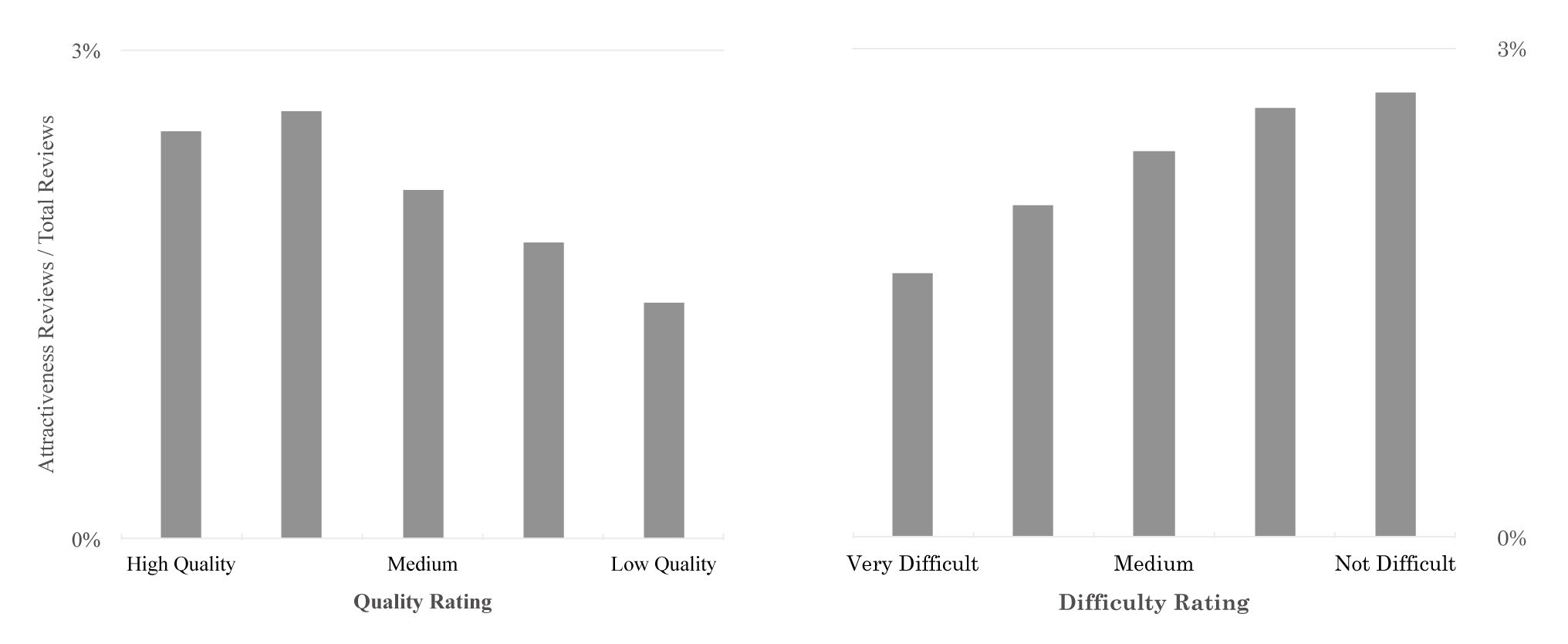}
\caption{Proportion of reviews with attractiveness commentary by quality and difficulty ratings.}
\label{fig:quality}

\end{figure*}

\paragraph{Quality and teacher gender}
We compare the proportions for attractiveness commentary in relation to quality and difficulty rating scales (\autoref{fig:quality}). There is a significant interaction between teacher quality and gender, female professors rated high quality are significantly more likely to receive attractiveness commentary than male professors rated high quality (see \autoref{tab:GEEModel}). Difficulty was also a significant factor, the higher the difficulty score, the less likely the reviews for the professor will contain attractiveness commentary.  

\section{Discussion}
While our work has focused on the text contents of reviews, our analysis of objectifying comments follows previous findings about biases of the original chili pepper rating, correlating with teacher gender, quality, and difficulty ratings. This is the first study to find a correlation between  attractiveness commentary and the website interface.

More research is needed to understand the observed steady eight-year decline.
As this was an observational study rather than a controlled experiment, there are many uncontrolled variables. For instance, we cannot compare attractiveness commentary by size of professor's class or attributes of the reviewer. We tried to estimate these factors with proxies such as university size, geographic area, and tuition amounts, but these only provide rough estimates and did not have significant effect on the presence of attractiveness commentary.
\citet{mcneil_lurking_2020} reflects on how users' perceptions of anonymity have changed, from posting to online bulletin boards in the late 1990s, to present-day \say{sharing} on corporate-owned, heavily surveilled social network sites like Facebook.  This turn from anonymity to self-awareness is observed by \citet{boyd}
in their study of Twitter users.
These users describe their own self-censoring behaviors by imagining their audiences to include not only friends but also parents and employers. The decline in attractiveness commentary on RMP may reflect broader internet trends, corresponding with internet users being more conscious of their perceived audience and realizing that true online anonymity is impossible.

\section{Conclusion}
We find that a small change to the RMP website, removal of the chili pepper rating, is associated with a lower likelihood of  comments on professor attractiveness. Our experiments show that an ensemble of classifiers can accurately detect objectifying language in online professor reviews and can allow us to analyze trends in a large unlabeled dataset.

One area where classifiers disagreed was in the \say{fuzzy samples} such as accents and godliness discussed in Section \ref{fuzz}. \citet{breitfeller-etal-2019-finding} describe similar challenges in classifying microaggressions and label themes within their dataset to better define these utterances. Based on our classifier's success in pulling out objectifying comments from large datasets, we can identify enough examples to consider labeling categories
such as accent criticism and comments about unattractiveness.
Finally, one could apply an active learning approach \citep{yarowsky-1995-unsupervised} to label and train on examples where the classifiers disagreed.

With further exploration it is hoped these techniques could be applied to detecting other forms of abusive language in online reviews. Insofar as the removal of the chili pepper feature correlated with a significant decrease in attractiveness commentary, we suggest that web interface design may positively influence online discourse. As the scope of the gig economy continues to expand and more workers find themselves evaluated by anonymous online reviews, we hope these findings will inspire future research around potential biases in online reviews based on gender, appearance, and the design of the online interface used. 

\section{Acknowledgments}
We would like to thank Martin Chodorow for his guidance in statistical analysis and Deepali Advani for her assistance with data preparation. We apprciate Jonathan Butterick for helping with data collection.
We would also like to acknowledge Sara Morini for her assistance with the data annotation,  William Jordan for proofreading, and anonymous reviewers for their helpful feedback.

\bibliographystyle{acl_natbib}
\bibliography{ghostpeppers}

\begin{thebibliography}{36}
\expandafter\ifx\csname natexlab\endcsname\relax\def\natexlab#1{#1}\fi

\bibitem[{Bird et~al.(2009)Bird, Klein, and Loper}]{bird_natural_2009}
Steven Bird, Ewan Klein, and Edward Loper. 2009.
\newblock \href {https://www.nltk.org/} {\emph{Natural language processing with
  {Python}}}.
\newblock O'Reilly.

\bibitem[{Boring(2017)}]{BORING201727}
Anne Boring. 2017.
\newblock \href
  {http://www.sciencedirect.com/science/article/pii/S0047272716301591} {Gender
  biases in student evaluations of teaching}.
\newblock \emph{Journal of Public Economics}, 145:27--41.

\bibitem[{Boring et~al.(2016)Boring, Ottoboni, and Stark}]{effectiveness}
Anne Boring, Kellie Ottoboni, and Philip~B. Stark. 2016.
\newblock \href
  {https://www.scienceopen.com/document?vid=818d8ec0-5908-47d8-86b4-5dc38f04b23e}
  {Student evaluations of teaching (mostly) do not measure teaching
  effectiveness}.
\newblock \emph{ScienceOpen Research}, 1:1--11.

\bibitem[{Breitfeller et~al.(2019)Breitfeller, Ahn, Jurgens, and
  Tsvetkov}]{breitfeller-etal-2019-finding}
Luke Breitfeller, Emily Ahn, David Jurgens, and Yulia Tsvetkov. 2019.
\newblock \href {https://www.aclweb.org/anthology/D19-1176} {Finding
  microaggressions in the wild: A case for locating elusive phenomena in social
  media posts}.
\newblock In \emph{Proceedings of the 2019 Conference on Empirical Methods in
  Natural Language Processing and the 9th International Joint Conference on
  Natural Language Processing (EMNLP-IJCNLP)}, pages 1664--1674.

\bibitem[{Brill and Wu(1998)}]{brill-wu-1998-classifier}
Eric Brill and Jun Wu. 1998.
\newblock \href {https://www.aclweb.org/anthology/P98-1029} {Classifier
  combination for improved lexical disambiguation}.
\newblock In \emph{36th Annual Meeting of the Association for Computational
  Linguistics and 17th International Conference on Computational Linguistics,
  Volume 1}, pages 191--195.

\bibitem[{Brock(2018)}]{CTDA}
André Brock. 2018.
\newblock \href {https://doi.org/10.1177/1461444816677532} {Critical
  technocultural discourse analysis}.
\newblock \emph{New Media \& Society}, 20(3):1012--1030.

\bibitem[{Chang and McKeown(2019)}]{chang-mckeown-2019-automatically}
Serina Chang and Kathy McKeown. 2019.
\newblock \href {https://www.aclweb.org/anthology/D19-1579} {Automatically
  inferring gender associations from language}.
\newblock In \emph{Proceedings of the 2019 Conference on Empirical Methods in
  Natural Language Processing and the 9th International Joint Conference on
  Natural Language Processing}, pages 5746--5752.

\bibitem[{Davison and Price(2009)}]{davison}
Elizabeth Davison and Jammie Price. 2009.
\newblock \href {https://doi.org/10.1080/02602930801895695} {How do we rate?
  {An} evaluation of online student evaluations}.
\newblock \emph{Assessment \& Evaluation in Higher Education}, 34(1):51--65.

\bibitem[{Felton et~al.(2008)Felton, Koper, Mitchell, and Stinson}]{attractive}
James Felton, Peter~T. Koper, John Mitchell, and Michael Stinson. 2008.
\newblock \href {https://doi.org/10.1080/02602930601122803} {Attractiveness,
  easiness and other issues: student evaluations of professors on
  ratemyprofessors.com}.
\newblock \emph{Assessment \& Evaluation in Higher Education}, 33(1):45--61.

\bibitem[{Flaherty(2018)}]{nochili}
Colleen Flaherty. 2018.
\newblock \href
  {https://www.insidehighered.com/news/2018/07/02/rate-my-professors-ditches-its-chili-pepper-hotness-quotient}
  {Bye, bye, chili pepper: {Rate My Professors} ditches its chili pepper
  ``hotness'' quotient}.
\newblock Accessed 10/28/2018.

\bibitem[{Freng and Webber(2009)}]{freng}
Scott Freng and David Webber. 2009.
\newblock \href {https://doi.org/10.1080/00986280902959739} {Turning up the
  heat on online teaching evaluations: Does ``hotness'' matter?}
\newblock \emph{Teaching of Psychology}, 36(3):189--193.

\bibitem[{van Halteren et~al.(1998)van Halteren, Zavrel, and
  Daelemans}]{van-halteren-etal-1998-improving}
Hans van Halteren, Jakub Zavrel, and Walter Daelemans. 1998.
\newblock \href {https://www.aclweb.org/anthology/P98-1081} {Improving data
  driven wordclass tagging by system combination}.
\newblock In \emph{36th Annual Meeting of the Association for Computational
  Linguistics and 17th International Conference on Computational Linguistics,
  Volume 1}, pages 491--497.

\bibitem[{Hempel(2017)}]{hempel_for_2017}
Jessi Hempel. 2017.
\newblock \href
  {https://www.wired.com/2017/02/for-nextdoor-eliminating-racism-is-no-quick-fix/}
  {For {Nextdoor}, {Eliminating} {Racism} {Is} {No} {Quick} {Fix}}.
\newblock \emph{Wired}.

\bibitem[{Herring(2004)}]{herring_2004}
Susan~C. Herring. 2004.
\newblock \href
  {https://www.cambridge.org/core/books/designing-for-virtual-communities-in-the-service-of-learning/ADDEEC1357AE22A19D69BAD552DBEB99}
  {Computer-mediated discourse analysis}.
\newblock In Sasha Barab, Rob Kling, and James~H.Editors Gray, editors,
  \emph{Designing for Virtual Communities in the Service of Learning}, page
  338–376. Cambridge University Press.

\bibitem[{Herring and Androutsopoulos(2015)}]{herring2}
Susan~C. Herring and Jannis Androutsopoulos. 2015.
\newblock \href {https://doi.org/10.1002/9781118584194.ch6} {Computer-mediated
  discourse 2.0}.
\newblock In Deborah Tannen, Heidi~E. Hamilton, and Deborah Schiffrin, editors,
  \emph{The Handbook of Discourse Analysis}, pages 127--151. John Wiley \&
  Sons.

\bibitem[{Honnibal and Montani(2017)}]{spacy2}
Matthew Honnibal and Ines Montani. 2017.
\newblock \href {https://spacy.io} {{spaCy 2}: natural language understanding
  with {Bloom} embeddings, convolutional neural networks and incremental
  parsing}.
\newblock Accessed 1/4/20.

\bibitem[{Kelly(2019)}]{Kelly_2019}
Bridget~Turner Kelly. 2019.
\newblock \href
  {https://www.brookings.edu/blog/brown-center-chalkboard/2019/03/29/though-more-women-are-on-college-campuses-climbing-the-professor-ladder-remains-a-challenge/}
  {Though more women are on college campuses, climbing the professor ladder
  remains a challenge}.

\bibitem[{Kindred and Mohammed(2017)}]{kindred}
Jeannette Kindred and Shaheed~N. Mohammed. 2017.
\newblock \href {https://doi.org/10.1111/j.1083-6101.2005.tb00257.x} {``he will
  crush you like an academic ninja!'': Exploring teacher ratings on
  {RateMyProfessors.com}}.
\newblock \emph{Journal of Computer-Mediated Communication}, 10(3).

\bibitem[{Kramer et~al.(2014)Kramer, Guillory, and Hancock}]{Kramer8788}
Adam D.~I. Kramer, Jamie~E. Guillory, and Jeffrey~T. Hancock. 2014.
\newblock \href {https://doi.org/10.1073/pnas.1320040111} {Experimental
  evidence of massive-scale emotional contagion through social networks}.
\newblock \emph{Proceedings of the National Academy of Sciences},
  111(24):8788--8790.

\bibitem[{Lagorio(2006)}]{Voice}
Christine Lagorio. 2006.
\newblock \href {https://www.villagevoice.com/2006/01/03/hot-for-teacher/} {Hot
  for teacher}.
\newblock \emph{The Village Voice}.

\bibitem[{Landis and Koch(1977)}]{landis_1977}
J.~Richard Landis and Gary~G. Koch. 1977.
\newblock \href {https://www.jstor.org/stable/2529310} {The measurement of
  observer agreement for categorical data}.
\newblock \emph{Biometrics}, 33(1):159--174.

\bibitem[{Liang and Zeger(1986)}]{GEE10.2307/2336267}
Kung-Yee Liang and Scott~L. Zeger. 1986.
\newblock \href {http://www.jstor.org/stable/2336267} {Longitudinal data
  analysis using generalized linear models}.
\newblock \emph{Biometrika}, 73(1):13--22.

\bibitem[{Marwick and boyd(2011)}]{boyd}
Alice~E. Marwick and danah boyd. 2011.
\newblock \href {https://doi.org/10.1177/1461444810365313} {I tweet honestly, i
  tweet passionately: Twitter users, context collapse, and the imagined
  audience}.
\newblock \emph{New Media \& Society}, 13(1):114--133.

\bibitem[{McLaughlin(2018)}]{byechili2}
BethAnn McLaughlin. 2018.
\newblock \href
  {https://edgeforscholars.org/i-killed-the-chili-pepper-on-rate-my-professor/}
  {I killed the chili pepper on {Rate My Professors}}.
\newblock Accessed 1/4/2020.

\bibitem[{{McNeil}(2020)}]{mcneil_lurking_2020}
Joanne {McNeil}. 2020.
\newblock \href {https://us.macmillan.com/lurking/joannemcneil/9780374194338}
  {\emph{Lurking: How a Person Became a User}}.
\newblock Macmillan.

\bibitem[{Pang and Lee(2004)}]{10.3115/1218955.1218990}
Bo~Pang and Lillian Lee. 2004.
\newblock \href {https://www.aclweb.org/anthology/P04-1035/} {A sentimental
  education: sentiment analysis using subjectivity summarization based on
  minimum cuts}.
\newblock In \emph{Proceedings of the 42nd Annual Meeting on Association for
  Computational Linguistics}, page 271–278.

\bibitem[{Pavlick and Tetreault(2016)}]{pavlick-tetreault-2016-empirical}
Ellie Pavlick and Joel Tetreault. 2016.
\newblock \href {https://www.aclweb.org/anthology/Q16-1005} {An empirical
  analysis of formality in online communication}.
\newblock \emph{Transactions of the Association for Computational Linguistics},
  4:61--74.

\bibitem[{Pedregosa et~al.(2011)Pedregosa, Varoquaux, Gramfort, Michel,
  Thirion, Grisel, Blondel, Prettenhofer, Weiss, Dubourg, Vanderplas, Passos,
  Cournapeau, Brucher, Perrot, and Duchesnay}]{scikit-learn}
Fabian Pedregosa, Gaël Varoquaux, Alexandre Gramfort, Vincent Michel, Bertrand
  Thirion, Olivier Grisel, Mathieu Blondel, Peter Prettenhofer, Ron Weiss,
  Vincent Dubourg, Jake Vanderplas, Alexandre Passos, David Cournapeau,
  Matthieu Brucher, Matthieu Perrot, and Édouard Duchesnay. 2011.
\newblock \href {http://www.jmlr.org/papers/v12/pedregosa11a.html}
  {Scikit-learn: machine learning in {Python}}.
\newblock \emph{Journal of Machine Learning Research}, 12:2825--2830.

\bibitem[{Ritter(2008)}]{Ritter}
Kelly Ritter. 2008.
\newblock \href {https://doi.org/10.1080/07350190802126177} {E-valuating
  learning: {Rate My Professors} and public rhetorics of pedagogy}.
\newblock \emph{Rhetoric Review}, 27(3):259--280.

\bibitem[{Rosen(2018)}]{rosen}
Andrew~S. Rosen. 2018.
\newblock \href {https://doi.org/10.1080/02602938.2016.1276155} {Correlations,
  trends and potential biases among publicly accessible web-based student
  evaluations of teaching: a large-scale study of ratemyprofessors.com data}.
\newblock \emph{Assessment \& Evaluation in Higher Education}, 43(1):31--44.

\bibitem[{Schmidt(2015)}]{schmidt}
Ben Schmidt. 2015.
\newblock \href {http://benschmidt.org/profGender/} {Gendered language in
  teacher reviews}.
\newblock Accessed 7/13/19.

\bibitem[{Statt(2020)}]{twitter_warning}
Nick Statt. 2020.
\newblock \href
  {https://www.theverge.com/2020/5/5/21248201/twitter-reply-warning-harmful-language-revise-tweet-moderation}
  {Twitter tests a warning message that tells users to rethink offensive
  replies}.
\newblock \emph{The Verge}.

\bibitem[{TextBlob(2018)}]{textblob}
TextBlob. 2018.
\newblock \href {https://textblob.readthedocs.io/en/dev/index.html}
  {{TextBlob}: simplified text processing}.
\newblock Accessed 1/8/2020.

\bibitem[{Tjong Kim~Sang and
  De~Meulder(2003)}]{tjong-kim-sang-de-meulder-2003-introduction}
Erik~F. Tjong Kim~Sang and Fien De~Meulder. 2003.
\newblock \href {https://www.aclweb.org/anthology/W03-0419} {Introduction to
  the {C}o{NLL}-2003 shared task: language-independent named entity
  recognition}.
\newblock In \emph{Proceedings of the Seventh Conference on Natural Language
  Learning at {HLT}-{NAACL} 2003}, pages 142--147.

\bibitem[{Wiebe et~al.(2001)Wiebe, Wilson, and Bell}]{wiebeetalacl01wkshop}
Janyce Wiebe, Theresa Wilson, and Matthew Bell. 2001.
\newblock \href
  {https://www.bibsonomy.org/bibtex/25b4444e30410fa3a0bde25953b830f8f/subjectivity}
  {Identifying collocations for recognizing opinions}.
\newblock In \emph{Proceedings of the ACL-01 Workshop on Collocation:
  Computational Extraction, Analysis, and Exploitation}, pages 24--31.

\bibitem[{Yarowsky(1995)}]{yarowsky-1995-unsupervised}
David Yarowsky. 1995.
\newblock \href {https://www.aclweb.org/anthology/P95-1026} {Unsupervised word
  sense disambiguation rivaling supervised methods}.
\newblock In \emph{33rd Annual Meeting of the Association for Computational
  Linguistics}, pages 189--196.

\end{thebibliography}
\clearpage

\appendix
\section{Appendix}
\label{sec:appendix}
\begin{figure}[ht!]
\centering
\includegraphics[width=140mm,scale=0.3]{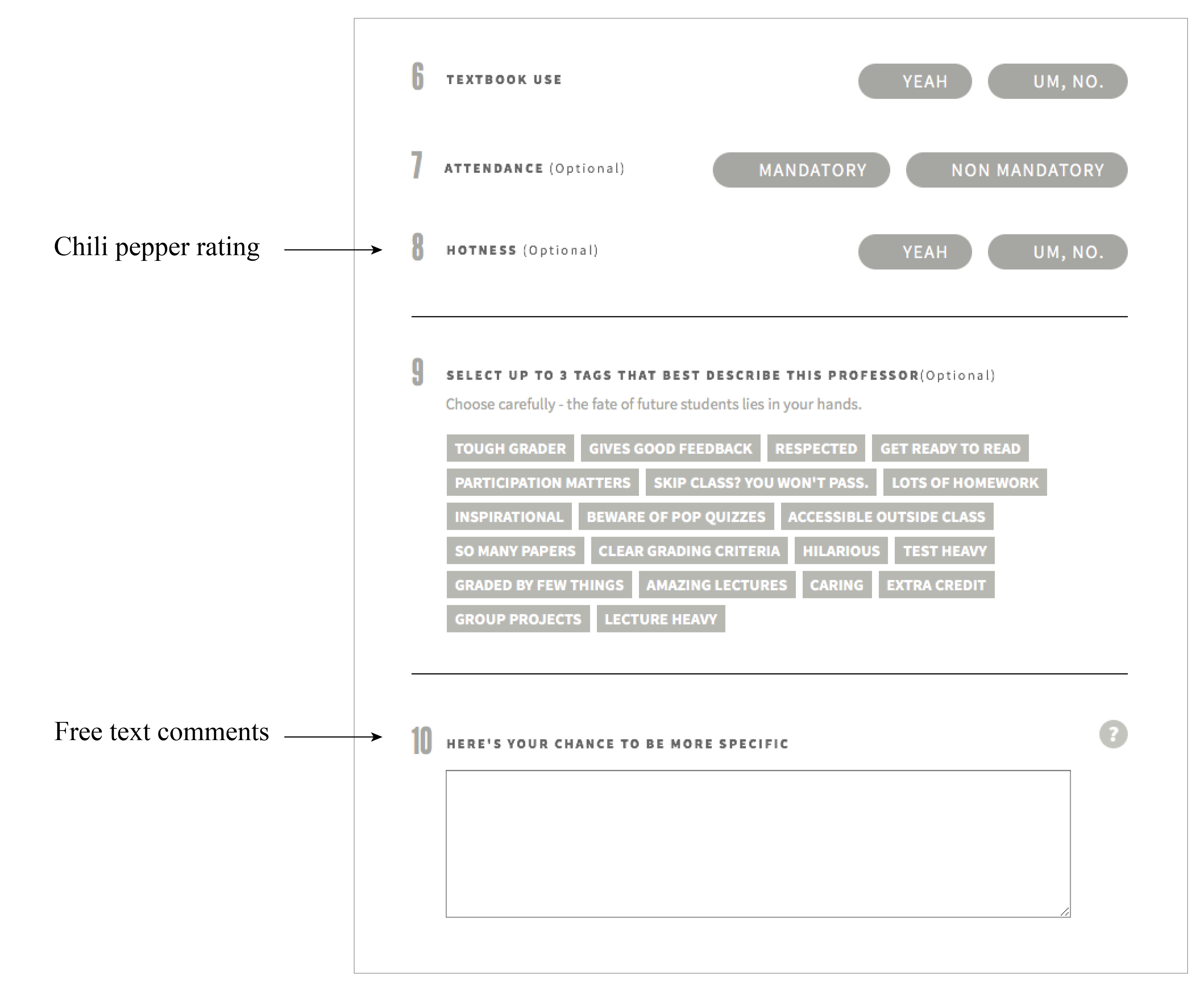}
\caption{Screen capture of review form on RateMyProfessors.com, 2014.}
\label{fig:webform}
\end{figure}

\begin{figure}[b!]
\centering
\includegraphics[width=140mm,scale=0.3]{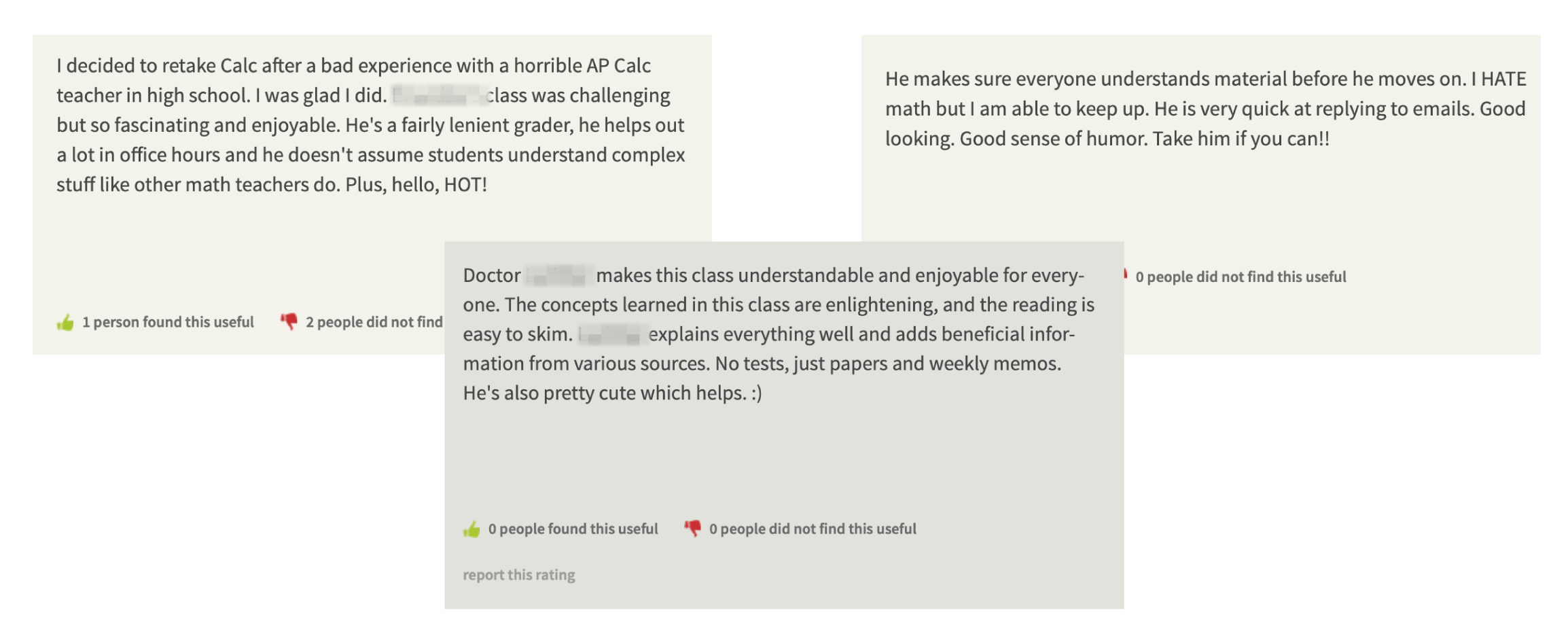}
\caption{Screen captures of attractiveness ratings on RateMyProfessors.com, 2019.}
\label{fig:GhostPeppers}
\end{figure}
\begin{figure*}[b!]
\centering
\includegraphics[width=140mm,scale=0.3]{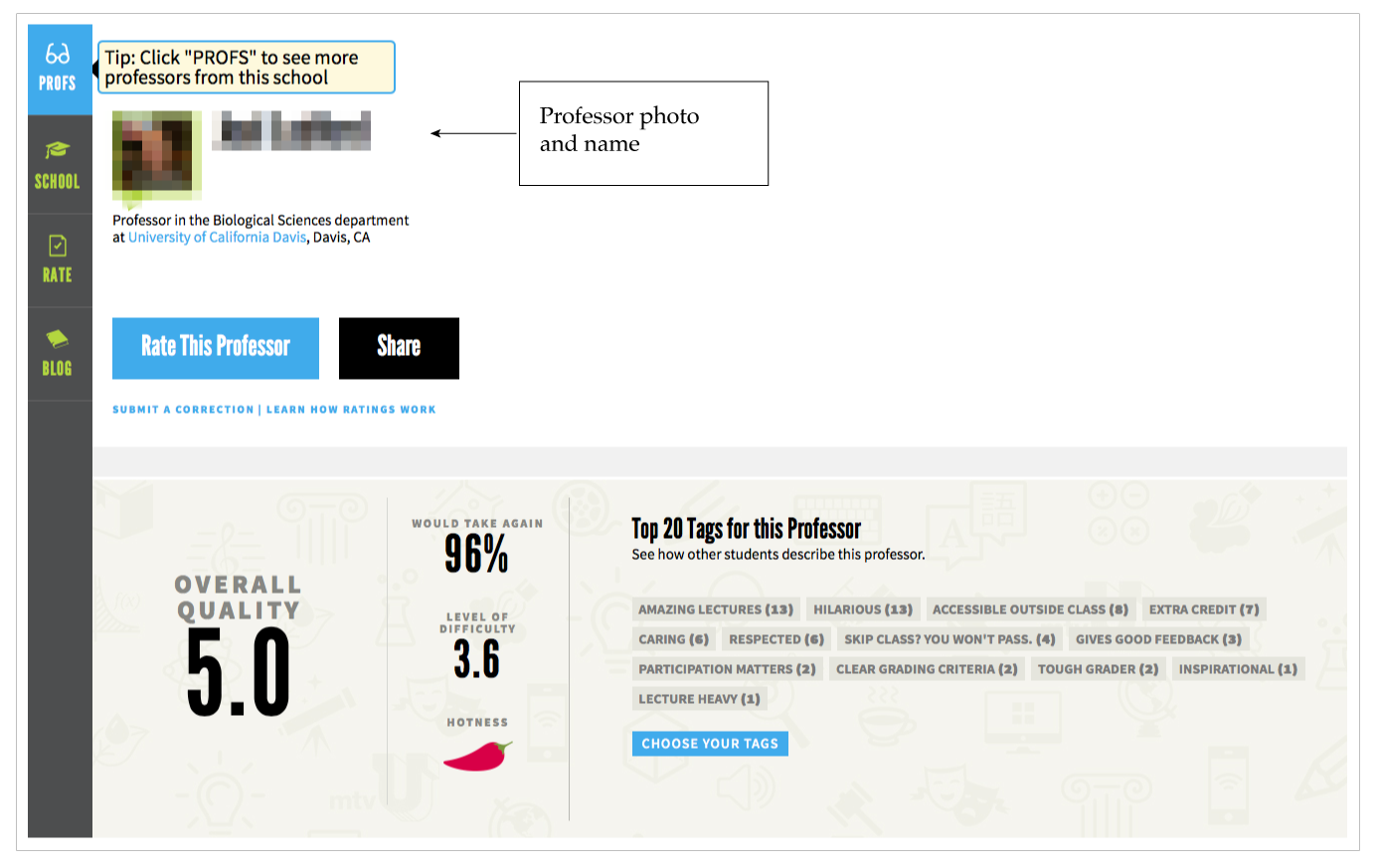}
\caption{Screen capture of professor profile on RateMyProfessors.com, 2014.}
\label{fig:prof_listing}
\end{figure*}
\begin{table*}
\small
\centering
\begin{tabular}{ll}
\toprule
Carnegie Mellon University & Stanford University\\
Duke University & Tufts University\\
Harvard University & University of Chicago \\
Massachusetts Institute of Technology & University of Texas at Austin\\
Princeton University & Yale University\\
Rice University \\
\bottomrule             
\end{tabular}
\caption{Universities sampled in labeled dataset.}
\label{table:d1colleges}   
\end{table*}
\begin{table*}[th!]\label{datsetuniv}
\centering
\small
\begin{tabular}{lll}
\toprule
Tuition Level & Enrollment Level & University\\
\midrule
high&low&Drexel University\\
&&Emory University\\
&&Fairfield University\\
&&Lawrence Technological University\\
&&Loyola Marymount University\\
&&Pennsylvania State University\\
&&Pomona College\\
&&Princeton University\\
&&Rice University\\
&&Trinity University\\
&&University of Chicago\\
&&University of Tulsa\\
&&Villanova University\\
&&Wesleyan University\\
&&Yale University\\
\midrule
high&medium& Northwestern University\\
&&Stanford University\\
\midrule
low&high&Iowa State University\\
&&University of California Los Angeles\\
&&University of South Carolina\\
&&University of Texas at Austin\\
&&University of Wisconsin\\
\midrule
low&low&College of Charleston\\
&&Evergreen State College\\
&&Montclair State University\\
&&New Mexico State University\\
&&Southern Utah University\\
&&University of Montana\\
&&University of Wyoming\\
\midrule
low&medium&Boise State University\\
&&Brigham Young University\\
&&Georgia Institute of Technology\\
&&Mississippi State University\\
&&Oklahoma State University\\
&&University of Illinois at Chicago\\
&&University of Northern Iowa\\
&&University of Oregon\\
&&Washington State University\\
&&Reed College\\
\midrule
medium&high&Rutgers State University\\
\midrule
medium&low&Austin College\\
&&Berry College\\
&&Bradley University\\
&&Newberry College\\
&&Oklahoma Baptist University\\
\midrule
medium&medium&Temple University\\

\bottomrule             
\end{tabular}
\caption{Universities appearing in unlabeled dataset. Tuition levels are binned by high (42,000--59,000), medium (27,000--41,000) and low (5,500--14,800). School enrollment is binned by high (35,000--52,000), medium (17,000--34,500) and low (2,000-17,000).}
\label{table:d2colleges}   
\end{table*}
\newpage
\begin{table*}[th!]
\centering
\begin{tabular}{p{0.25\textwidth}p{0.60\textwidth}}
\toprule
Dictionary&Word List\\
\midrule
hot dictionary&adorable, alluring, appealing, athletic, attractive, babe, bangin, banging, beaut, beautiful, beauty, becoming, beguiling, bewitched, bewitching, bootylicious, breathtaking, buxom, charming, chili, comely, cute, dainty, dazzling, divine, doll, dork, dorky, dreamboat, dreamy, enchanting, fetching, fire, flaming, fox, foxy, gentle, gentleness, glamorous, glorious, gorgeous, graceful, handsome, hottie, hubba, hunk,  hunky, hypnotic, irresistible, looker, lovely, luscious, magnetic, marry, nerdy, ravishing, seductive, sensuous, sexy, smokin, smoking, soothing, spiffy, striking, stunning, sublime\\
\midrule
fashion list&boots, clothes, clothing, dress, dressed, dresses, fashion, hip, hipster, jacket, outfit, outfits, shoes, socks, stylish, wardrobe, wear, wears\\
\midrule
hair words&bald, baldness, baldspot, beard, blond, blonde, brunette, curly, dreadlocks, hair, haircut, moustache, mustache, shave, sideburns, toupee, wavy\\

\bottomrule             
\end{tabular}
\caption{Domain-specific word lists used in feature selection.}
\label{table:dictionary}   
\end{table*}

\begin{table*}
    \centering
    \begin{tabular}{l c c c c c c c c c c  }
        \toprule
         \textbf{Features} \\
         \midrule
         Familiarity and first person& $\bullet$& \textendash & $\bullet$ &$\bullet$ &\textendash& \textendash& \textendash& \textendash& \textendash \\
         \midrule
         Lexical: \\
         has \say{hot} &$\bullet$& \textendash & $\bullet$& $\bullet$& $\bullet$& $\bullet$ &$\bullet$& $\bullet$ & $\bullet$ \\
         has \say{accent}& $\bullet$ &\textendash & $\bullet$ &$\bullet$& $\bullet$ &$\bullet$ &$\bullet$& \textendash &$\bullet$ \\
         age, body part, clothing &$\bullet$ & \textendash & $\bullet$& $\bullet$& $\bullet$ &$\bullet$ &$\bullet$& \textendash& \textendash\\
         \midrule
         Readability& $\bullet$& \textendash&  \textendash& \textendash &\textendash& \textendash& \textendash& \textendash& \textendash\\
         \midrule
         Sentiment polarity &$\bullet$& -& $\bullet$& $\bullet$& $\bullet$& \textendash& \textendash& \textendash& \textendash\\
         \midrule
         Subjectivity score & $\bullet$& \textendash& $\bullet$& $\bullet$& $\bullet$ &\textendash& \textendash& \textendash& \textendash \\
         \midrule
         Formality & $\bullet$ &\textendash& $\bullet$& \textendash& \textendash& \textendash& \textendash& \textendash& \textendash\\
         \midrule
         Pronouns & $\bullet$& \textendash& $\bullet$& $\bullet$& \textendash& \textendash& \textendash& \textendash&  \textendash \\
         \midrule
         Internet Style & $\bullet$& \textendash& $\bullet$& $\bullet$& $\bullet$& $\bullet$& \textendash& \textendash& \textendash \\
         \midrule
          \textit{F}1  &.90 & .27&.90&.90&.90&.90&.90&.79&.90\\
         \bottomrule
    \end{tabular}
    \caption{Feature ablation study demonstrating the document classifier's reliance on domain-specific lexical features in development dataset.}
    \label{table:featurestest}
\end{table*}


\nocite{textblob} 
\end{document}